\documentclass[runningheads]{llncs}

\usepackage[symbol]{footmisc}
\usepackage[mobile]{eccv}
\usepackage{multirow}

\usepackage{eccvabbrv}

\usepackage{graphicx}
\usepackage{booktabs}

\usepackage[accsupp]{axessibility}  % Improves PDF readability for those with disabilities.

\usepackage[pagebackref,breaklinks,colorlinks,citecolor=eccvblue]{hyperref}
\usepackage{orcidlink}

\begin{document}

\title{ComboVerse: Compositional 3D
    Assets Creation Using Spatially-Aware Diffusion Guidance} 

\titlerunning{ComboVerse}

\author{Yongwei Chen\inst{1}\protect\footnotemark[1]\protect\footnotemark[2] \and
Tengfei Wang\inst{2}\protect\footnotemark[1] \and
Tong Wu\inst{3} \and\\
Xingang Pan\inst{1} \and
Kui Jia\inst{4} \and
Ziwei Liu\inst{1}
}

\authorrunning{Chen et al.}

\institute{S-Lab, Nanyang Technological University
\and Shanghai Artificial Intelligence Laboratory 
\and The Chinese University of Hong Kong  
\and The Chinese University of Hong Kong, Shenzhen\\
\url{https://cyw-3d.github.io/ComboVerse/}
}

\maketitle
\footnotetext[1]{Equal contribution.}
\footnotetext[2]{Work done when interning at Shanghai Artificial Intelligence Laboratory.}
\begin{abstract}
 Generating high-quality 3D assets from a given image is highly desirable in various applications such as AR/VR. Recent advances in single-image 3D generation explore feed-forward models that learn to infer the 3D model of an object without optimization. Though promising results have been achieved in single object generation, these methods often struggle to model complex 3D assets that inherently contain multiple objects. In this work, we present \textbf{ComboVerse}, a 3D generation framework that produces high-quality 3D assets with complex compositions by learning to combine multiple models.
\textbf{1)} We first perform an in-depth analysis of this ``multi-object gap'' from both model and data perspectives.
\textbf{2)} Next, with reconstructed 3D models of different objects, we seek to adjust their sizes, rotation angles, and locations to create a 3D asset that matches the given image. 
\textbf{3)} To automate this process, we apply spatially-aware score distillation sampling (SSDS) from pretrained diffusion models to guide the positioning of objects. 
Our proposed framework emphasizes spatial alignment of objects, compared with standard score distillation sampling, and thus achieves more accurate results.
Extensive experiments validate ComboVerse achieves clear improvements over existing methods in generating compositional 3D assets.

\end{abstract}

\begin{figure}
  % \makebox[\textwidth][c]{}
  \includegraphics[width=\textwidth]{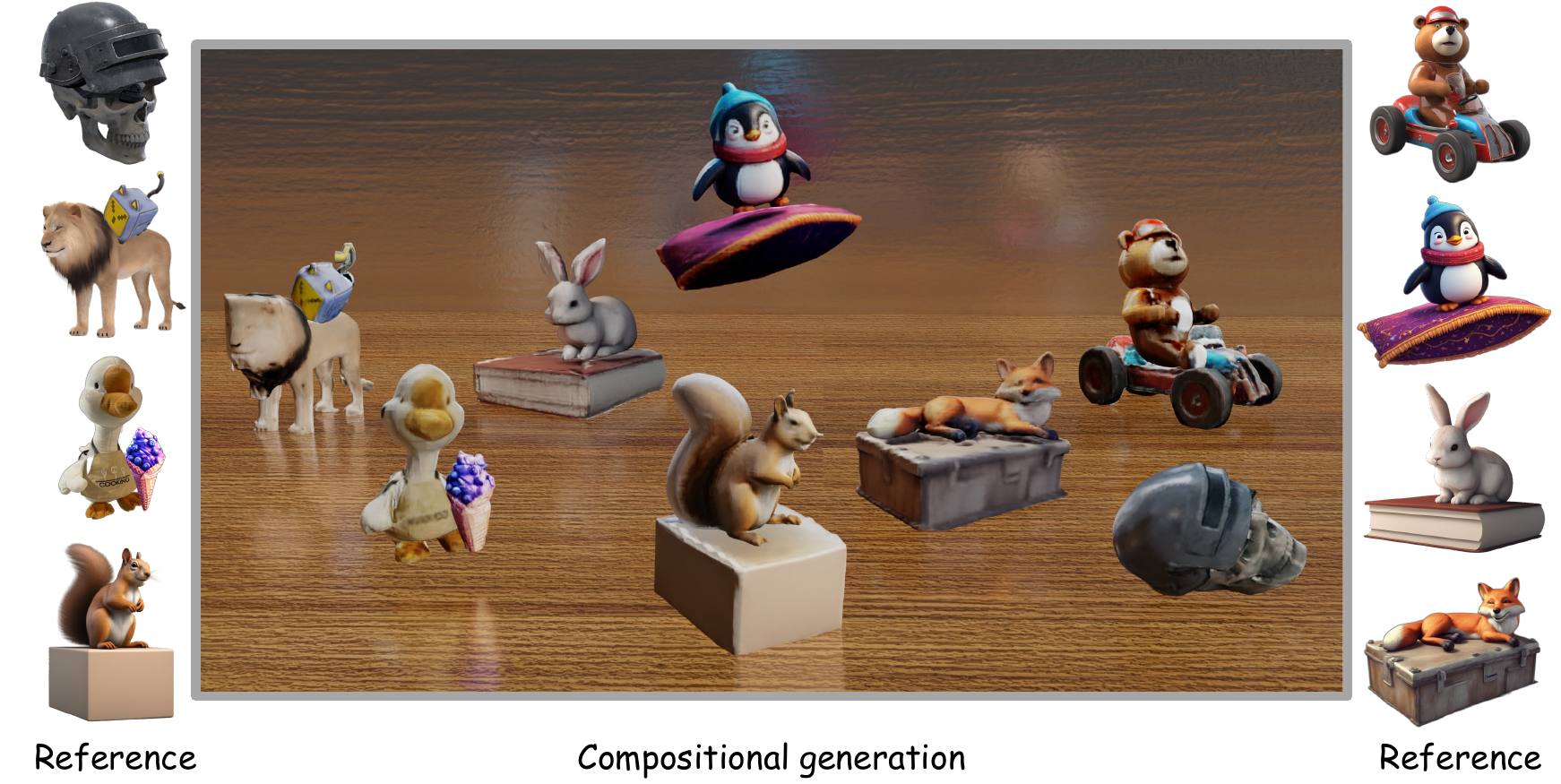}
  \caption{\textit{ComboVerse} can generate high-quality 3D models from a single image that contains multiple objects, \textit{e.g.,} a squirrel sitting on a paper box. We show textured meshes of created 3D content, showcasing stunning reconstruction quality.}
  \label{fig:teaser}
\end{figure}

\section{Introduction}
    Learning to create high-quality 3D assets from a single image is a long-standing goal in computer graphics and vision research, given its potential applications in AR/VR, movies, games, and industrial design. Over the years, a plethora of attempts have been made to leverage diffusion models~\cite{ho2020denoising} for 3D content creation.

Previously, limited 3D data availability led researchers to rely on pretrained 2D diffusion models for 3D guidance,  with a score distillation sampling (SDS) loss~\cite{poole2022_dreamfusion} transferring 3D-aware knowledge.  More recently, alternative approaches focus on training feed-forward 3D diffusion models for fast generation, facilitated by large-scale 3D object datasets like Objaverse~\cite{deitke2022_objaverse}. 
Once trained, these models can produce signed distance field~\cite{cheng2023_sdfusion}, points~\cite{nichol2022_pointe}, radiance fields~\cite{wang2022_rodin,jun2023_shape}, mesh~\cite{liu2023meshdiffusion}, or multi-view images~\cite{liu2023_zero1to3} through a single forward inference within one minute.

 Despite compelling results on simple object generation, these feed-forward methods usually encounter difficulties when applied to more complex data, such as scenes with multiple objects and complex occlusion. Fig.~\ref{fig:failure-analysis} illustrates the drawbacks of existing models when dealing with such combining objects.  However, upon generating each object separately, we observed that these models performed well. We perform an in-depth analysis of this ``multi-object gap'' and conjecture that this gap comes from the bias of their training data, \textit{i.e.,} Objaverse. The scarcity of 3D assets containing multiple objects makes it challenging for trained models to manage composites beyond the training data distribution.

Given the observations made above, is it possible to design a generative system that can produce 3D content containing multiple objects? Typically, skilled human artists create each object separately before integrating them into a whole. This has motivated us to present a compositional generation paradigm termed \textit{ComboVerse}, which generates each object individually and then focuses on automatically combining them to create a composite. A key advantage of our proposed paradigm is its ability to effectively manage complex assets containing multiple objects and occlusion.

Our approach comprises two stages: single-object reconstruction and multi-object combination. We first decompose and reconstruct each object within an image independently, using an occlusion removal module and an image-to-3D model. In the second stage, we aim to automatically combine the generated 3D objects into a single model, accounting for various factors such as object scale, placement, and occlusion. However, this process poses a challenge due to depth-size ambiguity in the input image, leading to inaccurate composition. 

To address this issue, we opt for pre-trained diffusion models as spatial guidance for object positioning.  Unlike previous SDS-based methods~\cite{poole2022_dreamfusion,lin2023_magic3d,chen2023_fantasia3d,tang2023_makeit3d} that require optimizing both the shape and texture from scratch, we fix the 3D model of individual objects and focus only on achieving a reasonable spatial layout so that the optimization process would be much faster. However, we have found that the standard SDS is insufficient for accurately placing objects, as it tends to prioritize content over position to match the given text prompt (see Fig.~\ref{fig:toy-examples}). To address this issue, we introduce a spatially-aware SDS loss that places greater emphasis on the spatial relationships between objects. Specifically, we reweight~\cite{hertz2022prompt} the attention map of the position tokens that indicate the spatial relation for score distillation. By prioritizing the awareness of position, the proposed loss can effectively distill the spatial knowledge from well-trained diffusion models for object placement.

To evaluate our method, we collect a benchmark consisting of 100 images that comprise a diverse range of complex scenes. We evaluate \textit{ComboVerse} on this benchmark, and extensive experiments show clear improvements over previous methods in terms of handling multiple objects, occlusion, and camera settings.  
% We also show various applications including object editing and part-level object assembly. 
Our main contributions can be summarized as:
\begin{itemize}
    \item We propose \textit{ComboVerse}, an automatic pipeline that extends object-level 3D generative models to generate compositional 3D assets from an image. 
    \item  We perform an in-depth analysis of the ``multi-object gap'' of existing feed-forward models from both model and data perspectives.    
    \item We propose spatially-aware diffusion guidance, enabling pre-trained image diffusion models to provide guidance on spatial layout for object placement.
    % \item  Extensive experiments show a clear improvement of the proposed method over previous approaches.
\end{itemize}

\section{Related works}
\textbf{3D Generation with 2D Diffusion Prior.}
Many methods opt to pretrained 2D diffusion models~\cite{ho2020denoising,rombach2022high} as a source of 3D guidance. Early works~\cite{poole2022_dreamfusion} proposed a score distillation sampling method to leverage the imaginative power of 2D diffusion for text-conditioned 3D content creation. Later works have improved the quality by using two-stage optimization~\cite{lin2023_magic3d,chen2023_fantasia3d,metzer2022_latentnerf}, better score distillation~\cite{wang2023_prolificdreamer}, and stronger foundation diffusion models~\cite{shi2023_MVDream,li2023_sweetdreamer}. Other works~\cite{xu2023_neurallift360,melaskyriazi2023_realfusion,tang2023_makeit3d,qian2023_magic123,sun2023dreamcraft3d,wang2023imagedream} extend the approach to generate 3D models from a single image. Some works~\cite{tang2023dreamgaussian,yi2023_gaussiandreamer} replace implicit representation with 3D gaussian splatting. Although the results are promising, creating a 3D model in this way can take several minutes to hours of optimization.

\noindent\textbf{Feed-forward 3D Generative Models.}
Another line of approaches trained feed-forward models for fast generation, eliminating the need for per-case optimization. 3D-aware generative adversarial networks~\cite{chan2021pigan,niemeyer2021giraffe,chan2022eg3d,or2022stylesdf,gu2021stylenerf,xiang2022gramhd} have gained considerable research interest in early research. Later, many attempts have been made to leverage diffusion models for image-conditioned and text-conditioned 3D generation. Once trained, they can produce signed distance field~\cite{cheng2023_sdfusion,chou2023_diffusionsdf}, points~\cite{nichol2022_pointe}, radiance fields~\cite{wang2022_rodin,gupta20233dgen,jun2023_shape,zhao2023michelangelo,hong20243dtopia}, mesh~\cite{liu2023meshdiffusion}, or multi-view images~\cite{liu2023_zero1to3,liu2023_syncdreamer,long2023wonder3d,liu2023one2345++,tang2024mvdiffusionpp} without optimization.
Besides diffusion models, recent works also explore feed-forward 3D reconstruction with transformer architecture~\cite{hong2023lrm,li2023instant3d} or UNet architecture~\cite{tang2024lgm}.
Despite fast generation, these methods are limited by the training data, restricting their ability to reconstruct complex 3D assets. We aim to build on these object-level generative models and extend them to handle more complex objects or scenes.

\noindent\textbf{Compositional 3D Generation.} Previous studies~\cite{niemeyer2021giraffe} have investigated the use of compositional neural radiance fields in an adversarial learning framework for the purpose of 3D-aware image generation. Additional studies have explored the concept of part-based shape generation, which involves assembling 3D parts into a 3D model. The seminal work~\cite{funkhouser2004modeling} retrieves from a mesh database to find parts of interest and composite the cut parts to produce novel objects. Later, the following works involve probabilistic models for part suggestion~\cite{kalogerakis2012probabilistic}, semantic attribute~\cite{chaudhuri2013attribit}, fabrication~\cite{Schulz:2014:FabByExample}, and CAD assembly~\cite{willis2022joinable}. Some works~\cite{tertikas2023partnerf} use neural radiance fields to represent different 3D components and then render these parts into a 3D model. With pretrained diffusion models as guidance,  recent work~\cite{po2023compositional} generates compositional 3D scenes with user-annotated 3D bounding boxes and text prompts. Concurrent works generate 3D scenes from text prompts by using large language models (LLMs) to propose 3D layouts as an alternative for human annotations~\cite{gao2023graphdreamer,vilesov2023cg3d,wang2023luciddreaming}, or jointly learning layout during optimization process~\cite{epstein2024disentangled}. These approaches can produce 3D scenes that match the text prompts, but texts can be unclear and imprecise when describing how objects are arranged in space. In contrast, our approach focuses on reconstructing complex 3D assets from a reference image. Unlike text descriptions that can be unclear or ambiguous, images represent the spatial relations among objects more accurately, requiring higher standards of composition quality.

\section{ComboVerse}

In this section, we will first analyze the ``multi-object gap'' of state-of-the-art image-to-3D generation methods trained on Objaverse, followed by a discussion of our compositional generative scheme. We then present the details of the two stages involved in our approach: single-object reconstruction and multi-object combination. The overview architecture is shown in Fig.~\ref{fig:overview}.

\begin{figure*}[t]
    \centering
    \includegraphics[width=\textwidth]{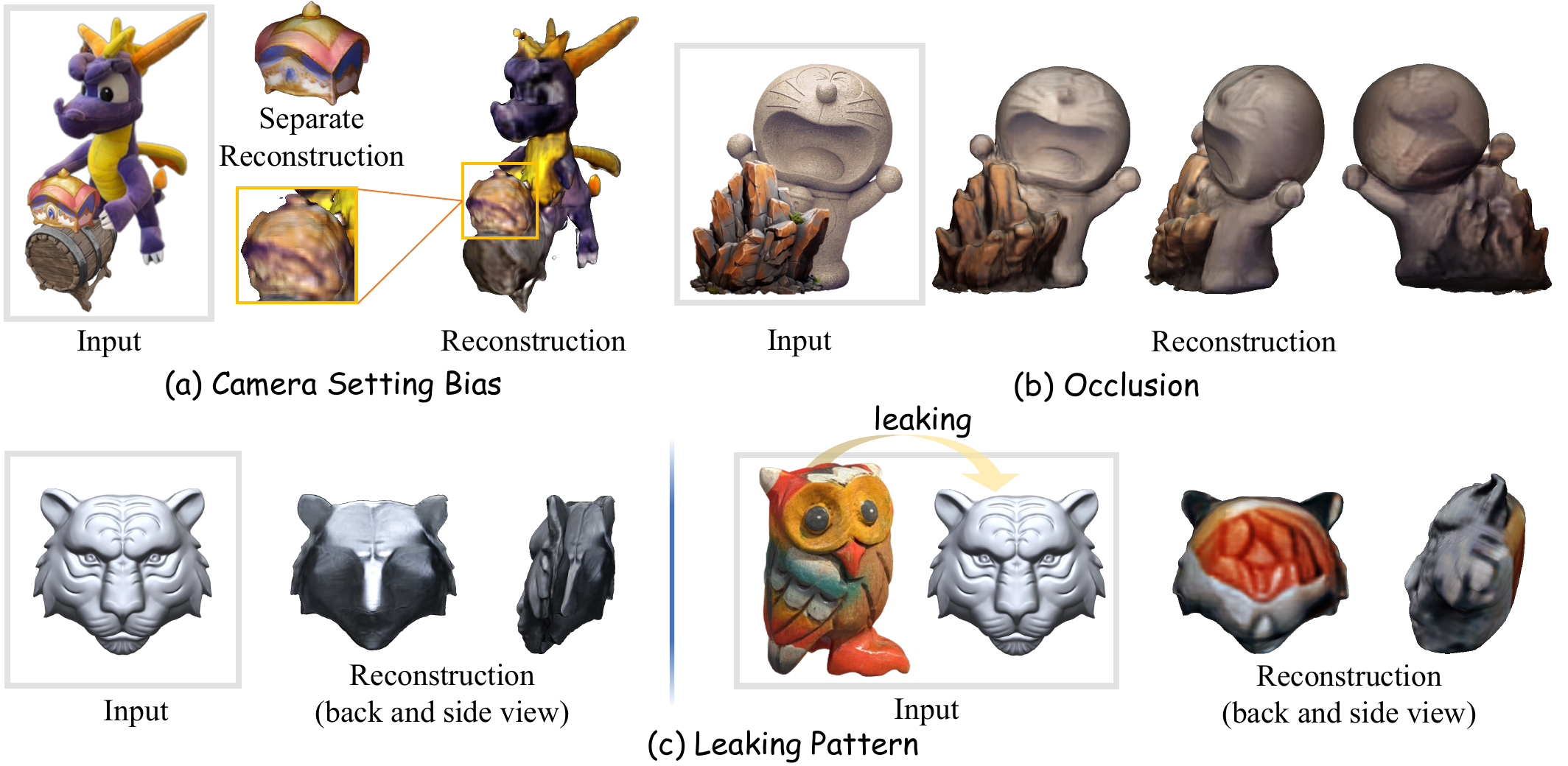}
    \caption{\textbf{``Multi-object gap'' of models trained on Objaverse.} (a) Camera Setting Bias. The reconstruction quality for small and non-centered objects will significantly downgrade compared to separate reconstruction.  (b) Occlusion. The reconstruction results tend to blend when an object is occluded by another. (c) Leaking Pattern. The shape and texture of an object will be influenced by other objects in the input image. For example, in (c), the tiger's back face adopts the owl's color, and its back surface becomes convex instead of concave due to the owl's shape influence.}
    \label{fig:failure-analysis}
\end{figure*}

\subsection{Analysis of ``Multi-Object Gap''}
Most existing feed-forward models are trained on Objaverse~\cite{deitke2022_objaverse}.
As shown in Fig.~\ref{fig:failure-analysis} and Fig.~\ref{fig:failure-analysis2}, these methods suffer three typical failure cases for multiple objects generation due to data and model biases.

\noindent\textbf{Camera Setting Bias.}
When setting up cameras, most image-to-3D methods assume that the object has a normalized size and is centered in the image. However, in scenarios with multiple objects, an object could appear in a corner of the scene or be very small in the image, which does not conform to object-centric assumptions. Such a case can result in a significant decline in modeling quality.

\noindent\textbf{Dataset Bias.} 
The Objaverse dataset predominantly features single-object assets, which poses a challenge for models trained on it to generalize to complex composites. Additionally, the near absence of occlusion in Objaverse results in these models struggling to handle occluded objects. As a result, generated objects often blend together due to occlusion ambiguity.

\noindent\textbf{Leaking Pattern.}
Existing methods tend to exhibit leakage issues when generating multiple objects simultaneously, where the geometry and appearance of one object can affect another. This issue may stem from the model's biases, as it is trained to generate a single object where different parts are consistent.  However, in scenes with multiple objects, different objects may have different geometry and texture. If they still affect each other, it can lead to bleeding patterns.

\noindent\textbf{Motivation.} As shown in Fig.~\ref{fig:failure-analysis}, though the current methods have difficulties in generating compositional objects, we have observed that these methods are successful in reconstructing each component object. This observation suggests the possibility of generating each object separately (Sec.~\ref{sec:3.2}) and subsequently combining them to form the desired compositional object (Sec.~\ref{sec:3.3}).

\begin{figure*}[t]
    \centering
    \includegraphics[width=\textwidth]{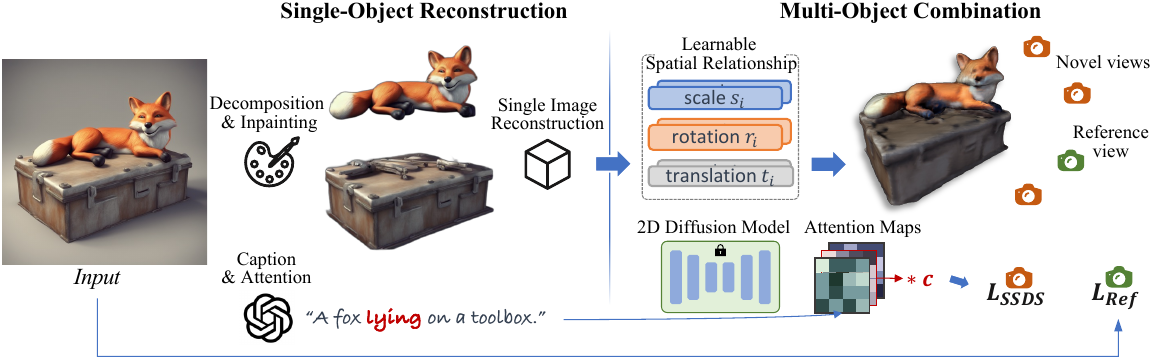}
    \caption{\textbf{Overview of our method.} Given an input image that contains multiple objects, our method can generate high-quality 3D assets through a two-stage process. In the single-object reconstruction stage, we decompose every single object in the image with object inpainting, and perform single-image reconstruction to create individual 3D models. In the multi-object combination stage, we maintain the geometry and texture of each object while optimizing their scale, rotation, and translation parameters $\{ s_i, r_i, t_i\}$. This optimization process is guided by our proposed spatially-aware SDS loss  $\mathcal{L}_{\mathrm{SSDS}}$, calculated on novel views, emphasizing the spatial token by enhancing its attention map weight. For example, considering the prompt \textit{``A fox lying on a toolbox.”} given to the 2D diffusion model, we emphasize the spatial token ``lying'' by multiplying its attention map with a constant $c$ ($c>1$). Also, we utilize the reference loss $\mathcal{L}_{\mathrm{Ref}}$, calculated on a reference view for additional constraints.}
    \label{fig:overview}
\end{figure*}

\subsection{Single-Object Reconstruction}
\label{sec:3.2}

\textbf{Components Decomposition.}
Given an input image \textit{I}, we first specify each object's 2D bounding box $\{ b_i \in \mathbb{Z}^4\}$, indicating the coordinates of the upper left and bottom right corners. 
Given bounding boxes $\{ b_i \in \mathbb{Z}^4\}$ for different objects, we use SAM~\cite{kirillov2023segany} to segment each object as follows: 
\begin{equation}
    O_i, M_i = \mathrm{SAM}(I, b_i),
\end{equation}
where $O_i$ and $M_i$ are the RGB channels and binary mask of $i$-th object.

\begin{figure}[t]
    \centering
    \begin{minipage}[t]{0.53\textwidth}
    \hspace{-1.5mm}\includegraphics[width=1.04\textwidth]{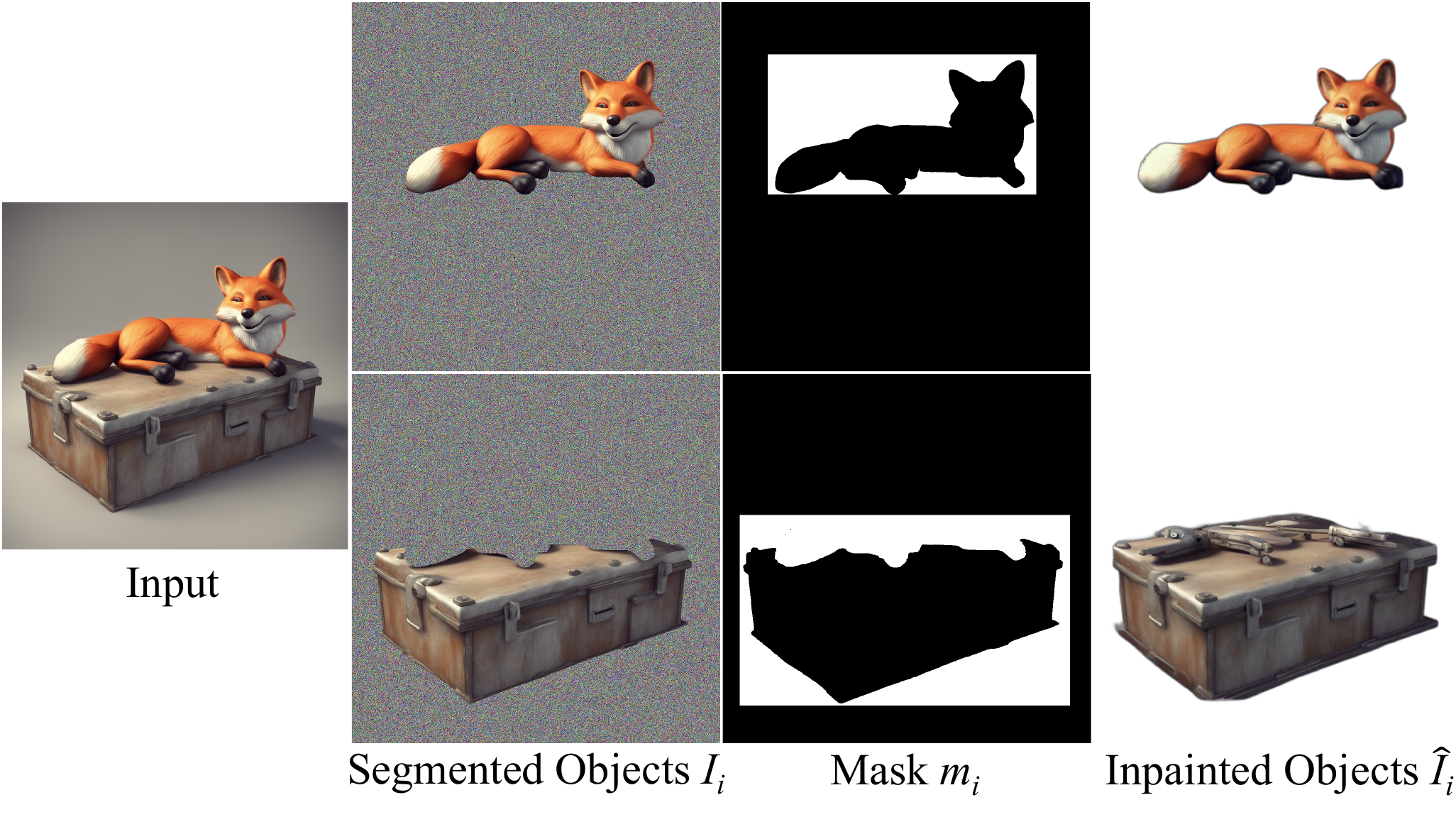}
    \caption{\textbf{Objects decomposition and inpainting.} In this stage, given an input image, we segment each separate object and get segmented objects with noise background image $I_{i}$ and bounding-aware mask $m_i$, then $I_{i}$ and $m_i$ are input to Stable Diffusion to obtain the inpainted objects $\hat{I}_i$.}
    \label{fig:component-proposal}
    \end{minipage}
    \hspace{3mm}
    \begin{minipage}[t]{0.42\textwidth}
    \hspace{2mm}
\includegraphics[width=0.93\textwidth]{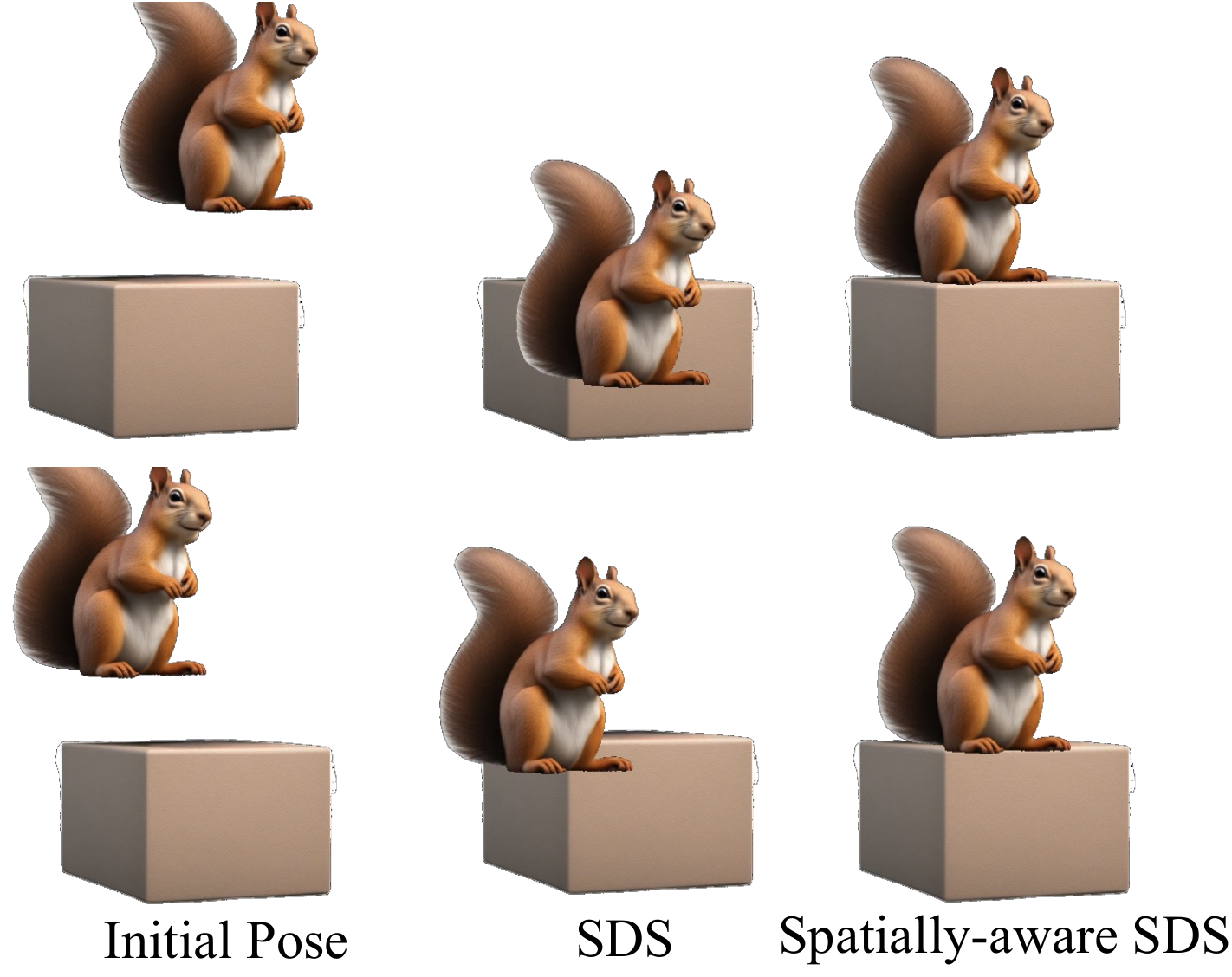}
    \caption{\textbf{2D toy examples.} We randomly initialize the squirrel with two different initial positions (left), and  optimize the position parameters to match the prompt ``a squirrel is sitting on a box''. Compared to standard SDS,  spatially-aware SDS produces better results.}
    \label{fig:toy-examples}
    \end{minipage}
\end{figure}

\noindent\textbf{Object Inpainting.}  To complete $O_i$ that is possibly occluded by another object, we utilize Stable Diffusion (SD)~\cite{rombach2022high} for objects inpainting.  However, we face the challenge of not having a known mask to identify occluded regions.  To address this issue, we design a strategy for completing the occluded parts of objects. First, to avoid generating a white or black border around the objects when inpainting, we replace the background of image $O_i$ with random noise, and the noised image $I_i$ is generated as follows:
\begin{equation}
    I_i = O_i + noise * (\sim M_i),
\end{equation}
where the $\sim M_i$ is the background region of $O_i$.
The noised image $I_i$ is illustrated in Fig. \ref{fig:component-proposal}.
Second, the background region and bounding box  $b_i$ are combined to generate an inpainting mask $m_i$ for each object, which indicates the inpainting region for each one. Specifically, for each inpainting mask $m_i$, the pixels that lie in the bounding box but outside the foreground object are set to 1, and the others are set to 0. That is:
\begin{equation}
    m_i = (\sim M_i) \cap b_i,
\end{equation}
where the $\sim M_i$ is the background region of $i$-th object $O_i$, and $b_i$ indicates its bounding box.
The mask $m_i$ is illustrated in Fig.~  \ref{fig:component-proposal}.
Finally, we input $I_i$ and $m_i$ that contain bounding box information to SD, to complete the object $I_i$ inside the bounding box $b_i$:  $\hat{I_i} = SD(I_i, m_i)$, where $\hat{I_i}$ is the completed object, which is illustrated in Fig.~ \ref{fig:component-proposal}. For better completion, we input a text prompt \textit{``a complete 3D model''} to SD when inpainting. After that, each inpainted object $\hat{I_i}$ can be reconstructed by  image-to-3D methods to produce single 3D models.

\subsection{Multi-Object Combination}
\label{sec:3.3}
At this stage, we seek to combine separate 3D models by optimizing their scale, rotation, and translation parameters $\{ s_i, r_i, t_i\}$, such that they align with the input image \textit{I} and semantic spatial relationships. We begin by initializing each object's scale, rotation, and translation based on \textit{I}, and then refine them using the proposed spatially-aware diffusion priors and guidance from the reference image. We will first introduce a spatially-aware diffusion distillation scheme, followed by a discussion on its application for automatic object combinations.

\noindent\textbf{Spatially-Aware Diffusion Guidance.}
DreamFusion \cite{poole2022_dreamfusion} presents a method that optimizes 3D representations from textual descriptions, by employing a pre-trained 2D diffusion model. The subject is represented as a differentiable parameterization \cite{mildenhall2020nerf}, where a differentiable MLP renderer $g$ renders 2D images $x = g(\theta)$ from a neural radiance field parameterized as $\theta$. It leverages a diffusion model $\phi$  to provide a score function $\hat{\epsilon}_{\phi}(x_t;y,t)$, which predicts the sampled noise $\epsilon$ given the noisy image $x_t$, text-embedding $y$, and noise level $t$. This score function guides the direction of the gradient for updating the neural parameters $\theta$, and the gradient is calculated by Score Distillation Sampling (SDS):
\begin{equation}\label{EqnSDS}
    \bigtriangledown_\theta \mathcal{L}_{\text{SDS}}(\phi,x) = \mathbb{E}_{t,\epsilon} \left [ w(t)(\hat{\epsilon}_{\phi}(x_t;y,t) - \epsilon)\frac{\partial x}{\partial \theta} \right ],
\end{equation}
while $w(t)$ is a weighting function. 

However, we find that SDS is unstable for position adjustment in our case. We use a text prompt \textit{``a squirrel is sitting on a box''} and an image of a squirrel and a box as a toy example, and aim to test the ability of SDS to adjust the position of the image elements according to the text prompt. As shown in Fig.~\ref{fig:toy-examples}, SDS does not produce the correct placement, as the image content (squirrel and box) already matches the prompt and SDS does not push the adjustment of position. We thus propose spatially-aware SDS to emphasize the position adjustment when calculating the gradient.

Recall that in SDS, for each text embedding $y$ and time step $t$, we use a UNet to predict the noise  $\hat{\epsilon}_{\phi}(x_t;y,t)$. The features of noisy image $\phi(x_t)$ are projected to a query matrix $Q=\mathcal{F}_{Q}(\phi(x_t))$, and the textual embedding is projected to a key matrix $K=\mathcal{F}_{K}(y)$ and a value matrix $V=\mathcal{F}_{V}(y)$, via the learned linear projections $\mathcal{F}_{Q}$, $\mathcal{F}_{K}$ and $\mathcal{F}_{V}$. The attention maps are then calculated by:
\begin{equation}\label{EqnSDS}
    M = \mathrm{Softmax} \left ( \frac{QK^T}{\sqrt{d}}  \right ), 
\end{equation}
where the $M_{j}$ indicates the attention map of $j$-th token,  and d is the latent projection dimension of the keys and queries.

To prioritize the refinement of spatial layout, we strengthen the key token that describes the spatial relationship. The key token can be the word describing spatial relationships, such as ``front'', ``on,'' and ``below,'' or the word describing object interaction, such as ``riding'' and ``holding'', which can be extracted by LLMs or indicated by the user. For example, consider the prompt        \textit{``a squirrel is sitting on a box''}, we want to strengthen the effect of the word   \textit{``sitting on''}, which describes the relationship between the squirrel and paper box. To achieve this spatially-aware optimization, we scale the attention maps of the assigned tokens $j^\star$ with a constant c ($c>1$), similar to ~\cite{hertz2022prompt}, resulting in a stronger focus on the spatial relationship. The rest of the attention maps remain unchanged:

\begin{equation}
    M:= \begin{cases}
        c \cdot  M_j\quad & \mathrm{if} \quad j = j^\star \\
        M_j \quad &\mathrm{otherwise}.
    \end{cases} 
\end{equation}
The spatially-aware SDS loss (SSDS) can be formulated as:
\begin{equation}\label{SSDS}
    \bigtriangledown_\theta \mathcal{L}_{\text{SSDS}}(\phi^\star,x) = \mathbb{E}_{t,\epsilon} \left [ w(t)(\hat{\epsilon}_{\phi^\star}(x_t;y,t) - \epsilon)\frac{\partial x}{\partial \theta} \right ],
\end{equation}
where the $\hat{\epsilon}_{\phi^\star}(x_t;y,t)$ is the predicted noise calculated with the strengthened attention maps which focus on the spatial words. For timesteps, we sample  $t$ from a range with high noise levels, as these steps have a bigger impact on the spatial layout of a generated image.

\noindent\textbf{Combine the objects.}
% In this stage, we have separate 3D models of each object in  input image $I$, and we need to optimize each object's spatial parameters, including scale,
% rotation and translation $\left \{s_i , r_i , t_i \right \}$.
We begin with a coarse initialization of scale,
rotation and translation $\left \{s_i , r_i , t_i \right \}$ from the bounding box $b_i$ and estimated depth $d_i$. Specifically, the scale $s_i$ is decided by the ratio of the bounding box size and image size:
\begin{equation}
s_i = max\left \{ \frac{W_{b_i}}{W_I}, \frac{H_{b_i}}{H_I} \right \},
\end{equation}
where the $W_{b_i}$, $H_{b_i}$, $W_I$, $H_I$ are the width and height of bounding box $b_i$ and input image $I$ respectively. As for translation $t_i$, we use a monocular depth prediction model~\cite{Ranftl2021} to estimate the average depth $d_i$ for $i$-th object, which is set as z-dimension in $t_i$, and the x and y dimensions are initialized by the center coordinates of bounding box $b_i$ and image size. That is: 
\begin{equation}
    \begin{array}{c} 
  t_i = ( X_{b_i} - \frac{W_I}{2}, Y_{b_i} - \frac{H_I}{2}, d_i), 
    \\
    d_i = \mathrm{Average}(\mathrm{Depth}(O_i)),
\end{array}
\end{equation}
where $X_{b_i}$, $Y_{b_i}$ are the center coordinates of bounding box $b_i$, $d_i$ is the average depth of each pixel that lie in $i$-th object $O_i$. The rotation angles in three dimensions $r_i$ are initialized as $(0,0,0)$.

However, due to depth-size ambiguity in a single-view image,  the predicted depth can be inaccurate, leading to an unreasonable initialization of depth and size. To alleviate the single-view ambiguity, we refine spatial parameters $\left \{s_i , r_i , t_i \right \}$ with the proposed spatially-aware SDS loss (SSDS) as novel-view supervision. To stabilize the optimization, we also constrain the reconstruction error between reference-view rendering $\hat{I}$ in  and  input image $I$:
\begin{equation}
\mathcal{L}_{\mathrm{Ref}} = \lambda_\mathrm{RGB} \left | \hat{I}_\mathrm{RGB} - I_\mathrm{RGB} \right | +  \lambda_\mathrm{A} \left | \hat{I}_\mathrm{A} - I_\mathrm{A} \right |,
\end{equation}
where $\lambda_\mathrm{RGB}$ and $\lambda_\mathrm{A}$ are weights for RGB  and alpha channels. The total loss is  a weighted summation of $\mathcal{L}_{\mathrm{Ref}}$ and $\mathcal{L}_{\mathrm{SSDS}}$.

\section{Experiments}

\subsection{Implementation Details}
\label{imple_details}
We set the guidance scale to 7.5 and the number of inference steps to 30 when inpainting the image with Stable Diffusion. We use Pytorch3D~\cite{ravi2020pytorch3d} as our differentiable rendering engine. We randomly sample timestep $t$ between 800 and 900. We use Adam as our optimizer, and the learning rate of translation on the z dimension is 0.01, while the others are set to 0.001. The loss weight $\lambda_\mathrm{Ref}$, $\lambda_\mathrm{SSDS}$, $\lambda_\mathrm{RGB}$, $\lambda_\mathrm{A}$,  are set to 1, 1, 1,000, 1,000 respectively. We set the multiplier $c$ for attention map to 25. We downsample each separate mesh to 50,000 faces in the multi-object combination stage and render 10 views for each iteration.

\subsection{Main Results}
\label{evaluation}
\noindent\textbf{Benchmark.}
% To the best of our knowledge, we are the first method that focuses on compositional reconstruction problems in single-image reconstruction.
To evaluate our method, we built a test benchmark containing 100 images covering a variety of complex 3D assets. The benchmark includes 50 images generated with stable diffusion, and 50 images constructed from real images with PhotoShop.  Each image has a foreground mask, a set of bounding boxes for objects, and a text caption. We use GPT4 to propose text prompts and spatial tokens, followed by manual filtering. We will make this benchmark publicly available.

\noindent\textbf{Comparison Methods.} 
We compare our method with three state-of-the-art single-image reconstruction methods: 1) SyncDreamer~\cite{liu2023_syncdreamer}, which we implement using the official code. The textured meshes are derived from the NeuS~\cite{wang2021neus} representation. 2) LRM~\cite{hong2023lrm}, which we implement using the publicly available code~\cite{openlrm}. 3) Wonder3D~\cite{long2023wonder3d}, which we implement using the official code and also use as our base model for image-to-3D reconstruction.

\begin{figure*}[t]
    \centering
    \includegraphics[width=1\textwidth]{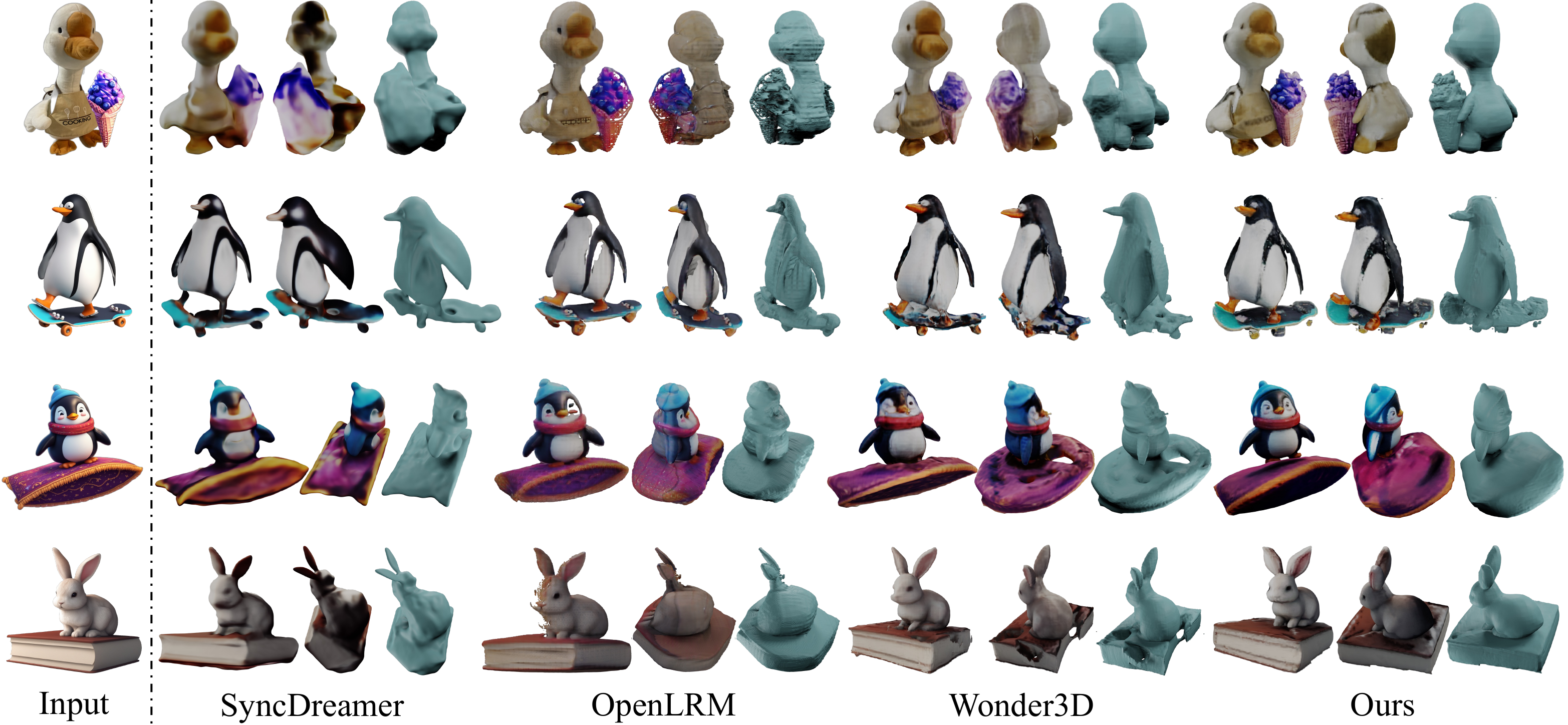}
    \caption{\textbf{Visual comparison for image-to-3D  generation.} Given an input image, previous methods reconstruct inaccurate geometry and blurry texture, especially in novel views. Our method produces higher-fidelity 3D models with the proposed compositional generation scheme. }
    \label{fig:comparison}
\end{figure*}

\begin{table}[t]
    \centering
    \caption{\textbf{Quantitative comparison.} }
    % \vspace{-10pt}
    \begin{tabular}{l @{\hspace{12mm}}c@{\hspace{8mm}} c}
    \toprule[1pt]
    Method  & CLIP-Score $\uparrow$   &  GPT-3DScore $\uparrow$\\
    \hline
    SyncDreamer \cite{liu2023_syncdreamer} & 81.47\% & 13.54\%\\
    OpenLRM \cite{hong2023lrm}   & 83.65\% &53.12\%\\
    Wonder3D \cite{long2023wonder3d}   & 85.57\%  & 56.25\%\\
    % \hline
    Ours  &  \textbf{86.58}\% & \textbf{65.63\%}\\
    \bottomrule[1pt]
    \end{tabular}
    \label{tab:quantitative_comparison}
\end{table}

\begin{figure*}
    \centering
    \includegraphics[width=1\textwidth]{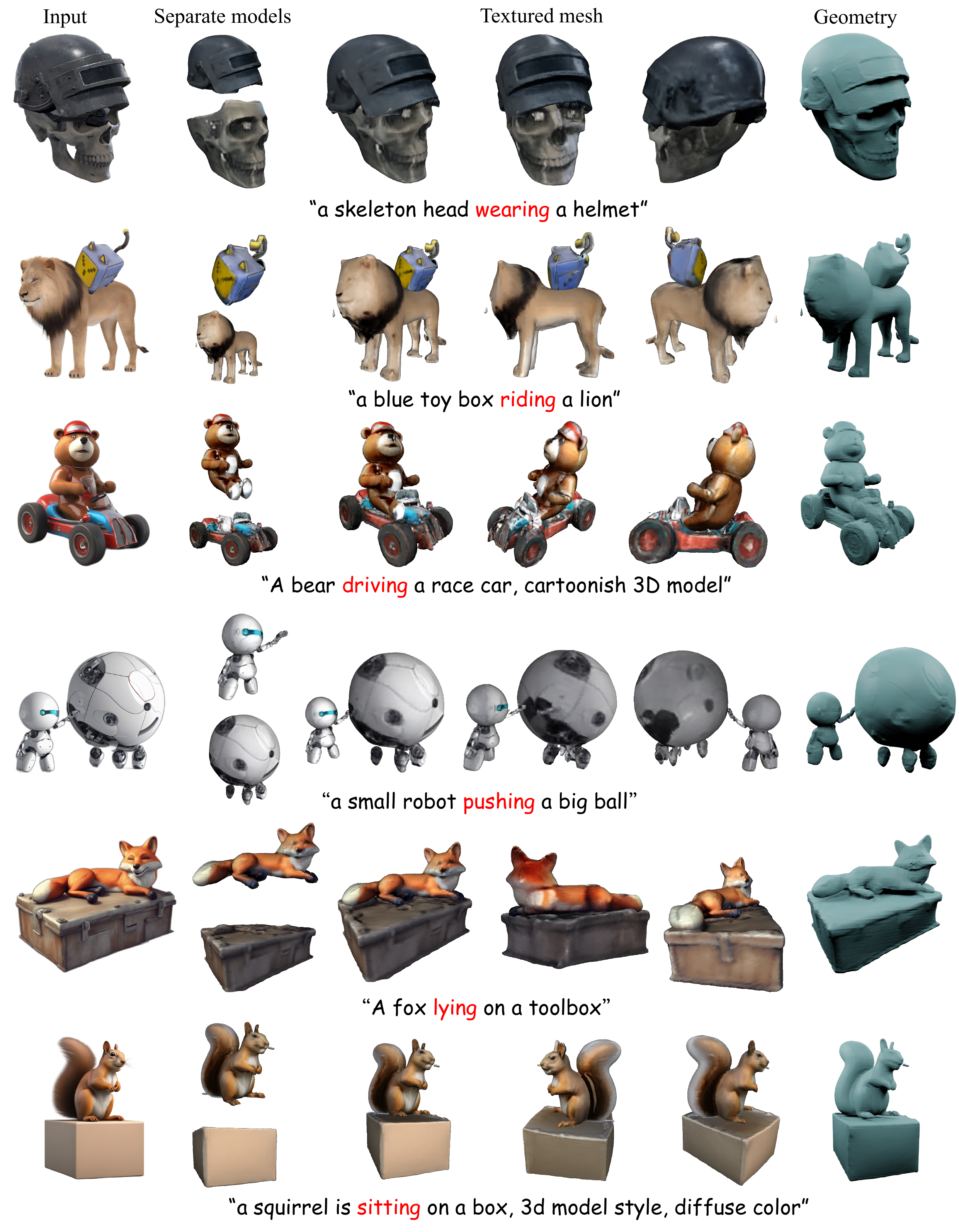}
    \caption{Qualitative results. \textit{ComboVerse} can generate high-quality 3D models from a single image that contains multiple objects.}
    \label{fig:results_supp}
\end{figure*}

\noindent\textbf{Qualitative Comparison.} 
As shown in Fig.~ \ref{fig:comparison}, our method can accurately reconstruct each object and preserve good spatial relationships among them. Other methods often struggle to generate high-quality geometry and texture for small objects in the input image, and the junction parts of different objects often blend. More qualitative results are shown in Fig.~\ref{fig:results_supp}.

\begin{figure}[t]
    \centering
    \includegraphics[width=0.95\textwidth]{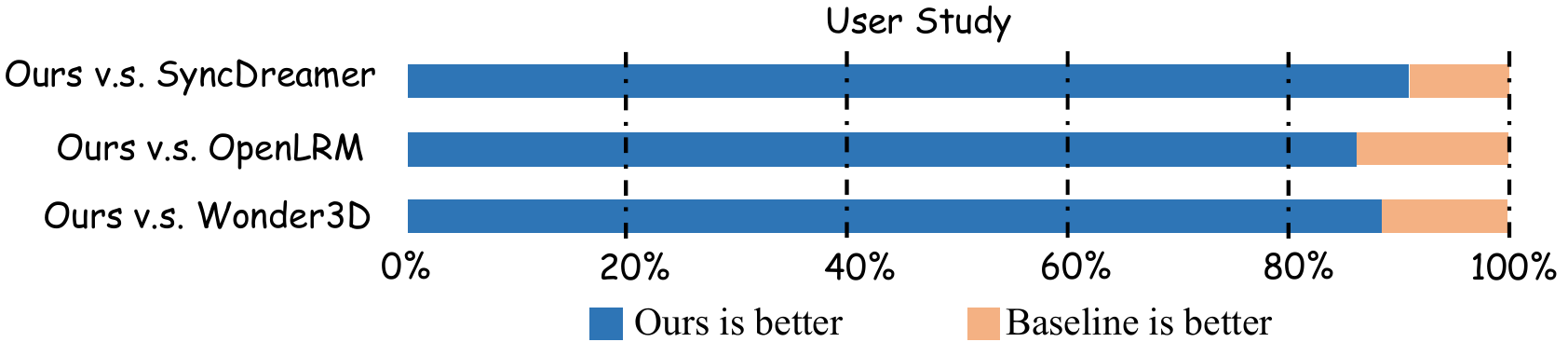}
    \caption{\textbf{User study.} Our method consistently outperforms competitors in terms of human evaluation. }
    \label{fig:user_study}
    \vspace{-8pt}
\end{figure}

\noindent\textbf{Quantitative Comparison.}
We use CLIP-Score~\cite{radford2021learning} to measure semantic similarities between novel-view images and the reference image.  We also involve GPT-based evaluation following~\cite{wu2023gpteval3d}. We conduct pair-wise comparisons for each method across all samples, and report the probability of success for each method. Table \ref{tab:quantitative_comparison} shows that our method outperforms comparison methods in both semantic similarity and GPT evaluation.

\noindent\textbf{User Study.}
Besides numerical metrics, we also perform a user study to compare our method with others. We collect 990 replies from 22 human users.  Participants are shown a reference image and a random pair of 3D models (ours and baselines) at once and are asked to select a more realistic one in terms of both geometry and texture quality. All choices are given in a shuffled order without time limitation.   Fig.~  \ref{fig:user_study} illustrates that our method outperforms previous approaches in terms of human preference.

\begin{figure}[t]
    \centering
    \includegraphics[width=0.95\textwidth]{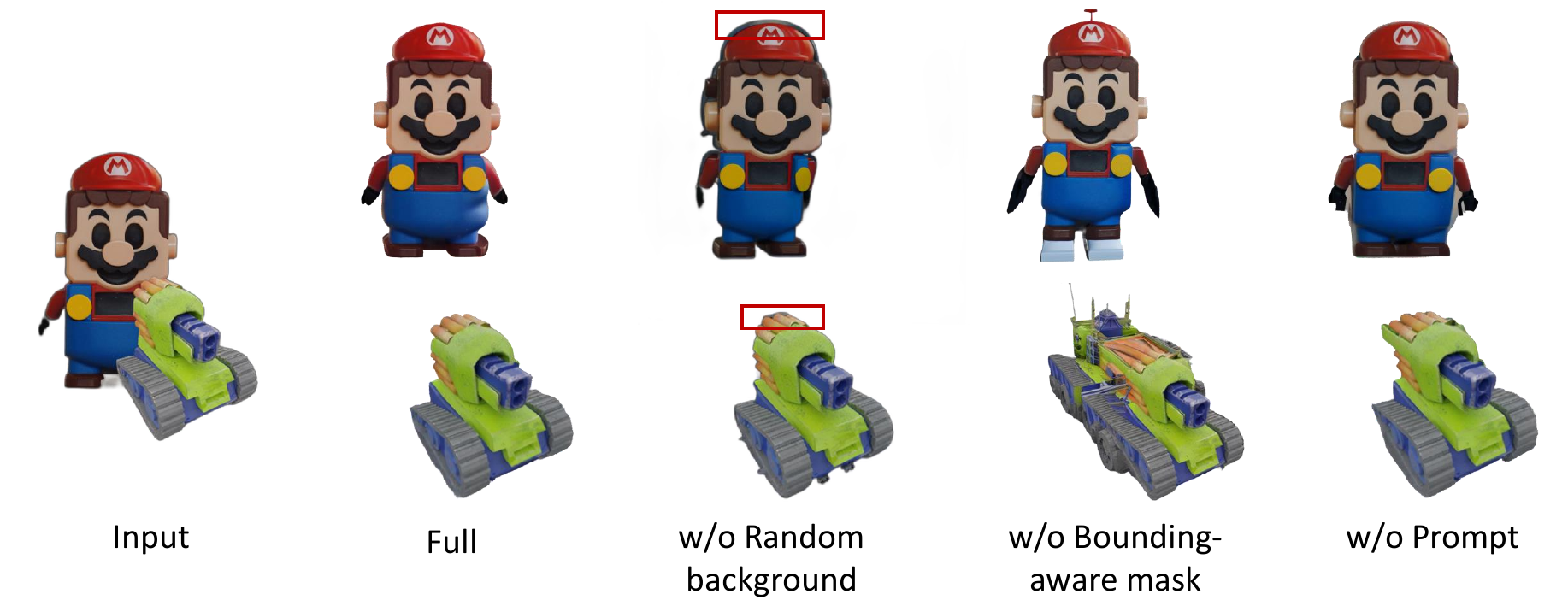}
    \caption{\textbf{Analysis for objects inpainting.} We produce compelling inpainting results with random background (EQ. 2), bounding-aware mask proposal (EQ. 3), and text prompting. }
    \label{fig:ablation image}
\end{figure}

\subsection{Ablation Study}
\label{ablation_study}

\noindent\textbf{Effectiveness of Object Inpainting.} We performed a study to examine designs in object completion. We first investigate the effect of noisy backgrounds on object completion.  The inpainted results without noise background had black borders and were not complete, as shown in Fig.~  \ref{fig:ablation image}. Then, we simply used the background mask for inpainting instead of a bounding-aware mask, and the inpainted results had extra parts, which caused inconsistency in the 3D reconstruction with the input image. We also removed the text prompt ``a complete 3D model'' for diffusion models during inpainting, and the results also degraded.

\begin{figure*}[t]
    \centering
    \includegraphics[width=\textwidth]{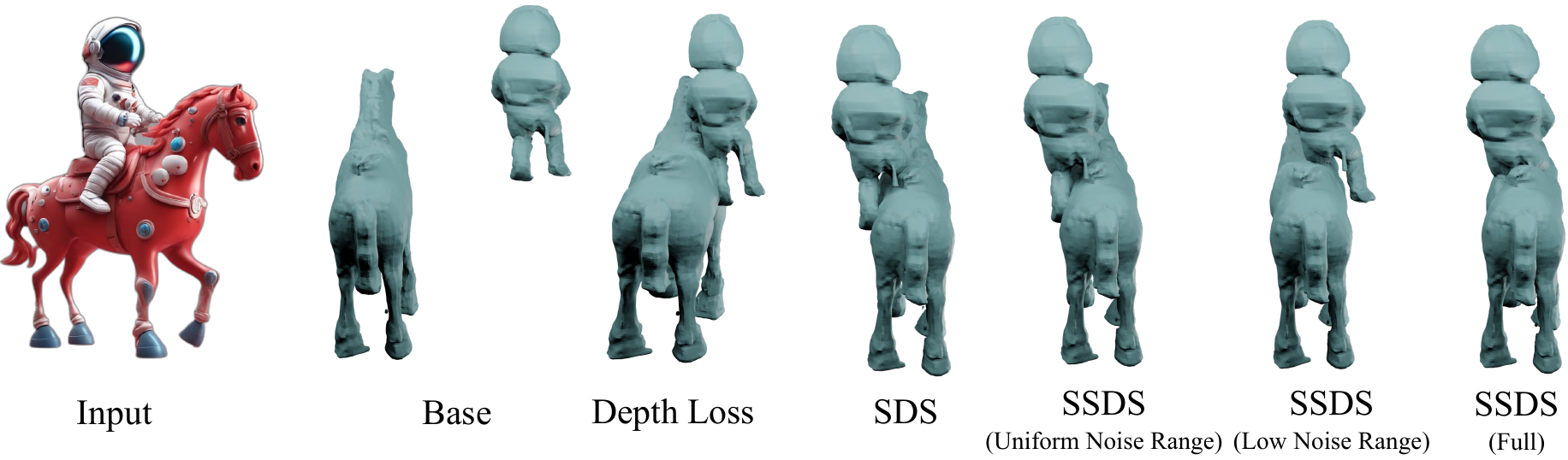}
    \caption{\textbf{Analysis for objects combination.} Compared with standard SDS and depth constrain, SSDS provides stronger guidance on object positioning. }
    \label{fig:ablation study}
    % \vspace{-12pt}
\end{figure*}

\noindent\textbf{Effectiveness of Spatially-Aware Diffusion Guidance.} 
As shown in Fig.~\ref{fig:ablation study}, we use an example of "an astronaut is riding a red horse"  to analyze different guidance settings in object combination.
\textit{Base} only enforces reconstruction loss in the reference view without additional guidance in novel views, and thus yields incorrect relative depth between the astronaut and the horse. With a standard \textit{SDS loss} or \textit{depth loss} from a depth prediction model as spatial guidance,   the interaction between the astronaut and the horse improves, but it was still far from accurate. By strengthening the attention to the word ``riding'' with the proposed \textit{SSDS loss (full)}, the full model achieves the best result.   This confirms the improved spatial control capability of the proposed method over the standard SDS. As discussed in Sec.~\ref{sec:3.3}, we sampled from a high noise range ([800, 900]) for Stable Diffusion when performing SSDS, as these steps have a bigger impact on the spatial layout of a generated image.  We also experiment with SSDS with different sample ranges of noise timesteps, \textit{low noise range} ([100, 200]), and \textit{uniform noise range} ([20, 980]) and observe a performance drop. We also give quantitative ablation results in \textit{supplementary materials.}.

\begin{figure}[t]
    \centering
    \includegraphics[width=0.95\textwidth]{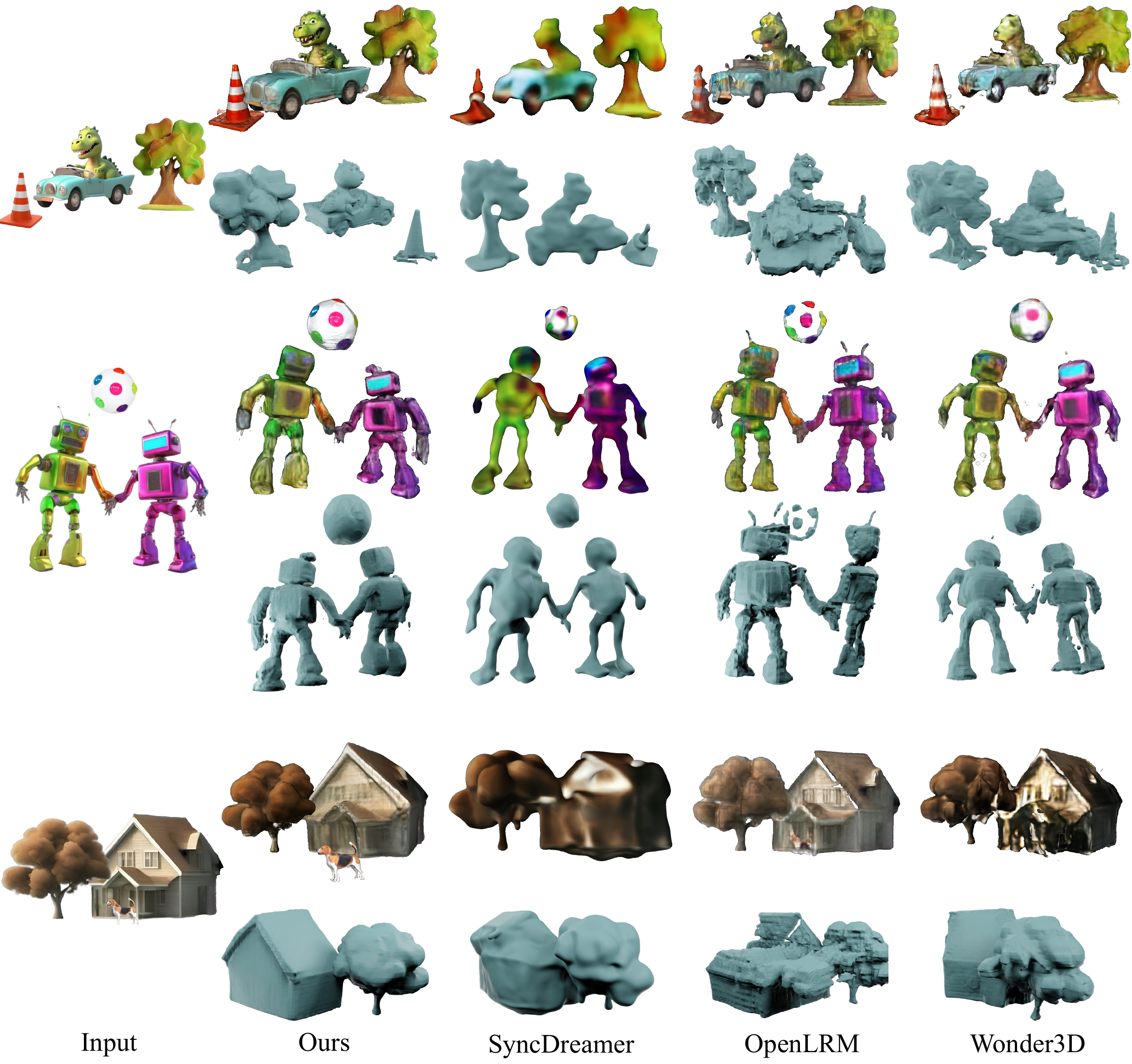}
    \caption{\textbf{Comparison of scene reconstruction.} We show some challenging cases that contain more than two objects. The first example involves four objects: \textit{a car, a dinosaur, a tree, and a cone}. The second example involves three examples: \textit{two robots and a ball}. The third example involves three examples: \textit{a house, a dog, and a tree}. Our method achieves compelling reconstruction quality with the compositional scheme. }
    \label{fig:scene}
\end{figure}

\subsection{Application in Scene Reconstruction}
\label{scene}
Besides the generation of 3D assets with two objects, we also validate the generalization ability of the proposed method to multiple objects. As illustrated in Fig.~ \ref{fig:scene}, we use the proposed method to reconstruct 3D scenes consisting of multiple ($>2$) objects.  Previous methods that work on the object level have difficulties in generating scenes and produce obvious artifacts in both geometry and texture. Also, for small objects such as the dog in the last example, existing methods tend to ignore it due to the training bias mentioned before. In contrast, our method achieves realistic and high-quality reconstruction.

\section{Conclusion}
In this paper, we present \textit{ComboVerse}, a novel method for creating high-quality compositional 3D assets from a single image. With an in-depth analysis of the ``multi-object gap'', we build on object-level 3D generative models and extend them to deal with more complex objects. With reconstructed 3D models of various objects, we seek to adjust their sizes, rotation angles, and locations to create a 3D asset that matches the given image. To this end, we proposed spatially-aware score distillation sampling from pretrained diffusion models to guide the placement of objects.  Our approach can be a valuable step for complex 3D object reconstruction and pave the way for future 3D scene generation.

\noindent\textbf{Limitations.} The proposed method performs well for assets consisting of two or a few (usually $<5$) objects. However, similar to existing text-based works, our method still faces challenges in creating very complex scenes with more objects.  Another limitation of our approach is the lack of optimization for the geometry and texture in the combination process. Thus, the quality of the final results relies on the performance of the image-to-3D method that we use as backbone. We expect that our methods can be further enhanced with more robust backbone methods in the future.

\section*{Appendix}
\appendix
\section{Quantitative Ablation Analysis}
To evaluate the effectiveness of the proposed SSDS, we performed an ablation analysis and have shown qualitative results in Fig. 10 in the main paper. Beyond visual comparison, we also provide quantitative ablation analysis in table \ref{tab:quantitative_comparison2}. Two CLIP models, including CLIP B/16 and ResNet50 backbones, are utilized to examine different guidance configurations in object combinations. To disentangle geometry and appearance quality, we evaluate multi-view CLIP similarities for both colored rendering and untextured geometry rendering. \textit{Base} only imposes reconstruction loss in the reference view, lacking additional guidance in novel views. Applying either  a standard 
 \textit{SDS} loss or \textit{depth} loss from a depth prediction model as spatial guidance yielded sub-optimal CLIP scores for color and geometry. However, by strengthening the attention to spatial layout through the 
 proposed \textit{SSDS loss (full)}, the full model achieves the best result, confirming its enhanced spatial control over standard SDS. As discussed in Sec. 3.3, high noise intervals ([800, 900]) were selected for Stable Diffusion during SSDS due to their bigger impact
 on the spatial layout of a generated image. We also experiment with SSDS with 
 different sample ranges of noise timesteps, \textit{low noise range} ([100, 200]), and 
 \textit{uniform noise range} ([20, 980]) and observe a performance drop. 

\begin{table}
    \centering
    \caption{\textbf{Quantitative analysis for ablation study.} }
    \begin{tabular}{l@{\hspace{8mm}}cccc}
    \toprule[1pt]
    \multirow{3}{*}{Guidance}  & \multicolumn{4}{c}{CLIP Score$\uparrow$}    \\
    & \multicolumn{2}{c}{CLIP B/16} & \multicolumn{2}{c}{ResNet50} \\
    & \makebox[0.13\textwidth][c]{Color} & \makebox[0.13\textwidth][c]{Geometry} & \makebox[0.13\textwidth][c]{Color} & \makebox[0.13\textwidth][c]{Geometry} \\
    \hline
    Base (without guidance) & 86.62\% & 75.24\% & 80.35\% & 74.19\%\\
    Depth Loss & 84.57\% & 78.42\% & 81.69\% & 75.83\%\\
    SDS & 84.16\% & 78.25\% & 84.08\% & 74.66\% \\
    % \hline
    SSDS (uniform noise range)   &  85.33\% & 78.49\% & 85.55\% & 75.85\%\\
    SSDS (low noise range)   &  84.86\% & 79.03\% & 84.42\% & 75.44\%\\
    SSDS (full)   &  \textbf{89.01}\% & \textbf{79.66}\% & \textbf{86.60}\% & \textbf{78.10}\%\\
    \bottomrule[1pt]
    \end{tabular}
    \label{tab:quantitative_comparison2}
\end{table}

\section{More Results of ``Multi-Object Gap''}
\begin{figure*}
    \centering
    \includegraphics[width=\textwidth]{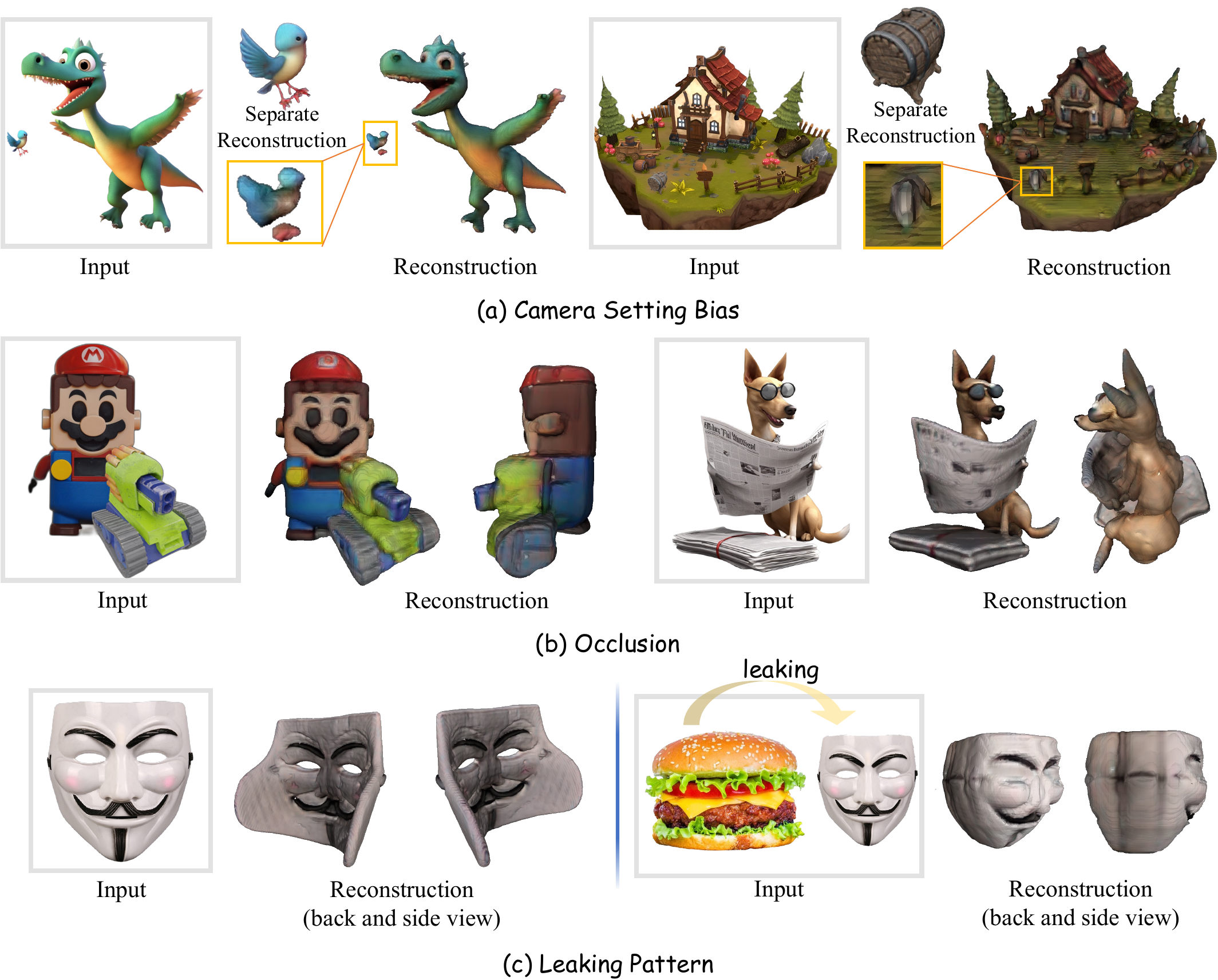}
    \caption{\textbf{``Multi-object gap'' of models trained on Objaverse.} (a) Camera Setting Bias. The reconstruction quality for small and non-centered objects will significantly downgrade compared to separate reconstruction. (b) Occlusion. The reconstruction results tend to blend when an object is occluded by another. (c) Leaking Pattern. The shape and texture of an object will be influenced by other objects in the input image.}
    \label{fig:failure-analysis2}
\end{figure*}

As discussed in Sec.3.1, the current feed-forward models, mainly trained on Objaverse, exhibit limitations in generalizing to multi-object scenarios.  Fig. \ref{fig:failure-analysis2} uses TripoSR, a state-of-the-art method in image-to-3D reconstruction, as another case to demonstrate this limitation.
Despite its advancements, TripoSR still exhibits three typical failure modes when tasked with generating multiple objects, stemming from inherent data and model biases. The detailed analysis was illustrated in Sec. 3.1.

\clearpage
\bibliographystyle{splncs04}
\bibliography{main}
\end{document}